\pdfoutput=1
\documentclass[leqno,11pt]{article}
\usepackage{arxiv_preprint}
\usepackage{graphicx}
\usepackage{microtype}

\runningauthors{Chen}
\runningtitle{How much does corpus choice change dependency-distance estimates?}

\title{How Much Does Corpus Choice Change Dependency-Distance Estimates?}

\addauthor{1*}{Sirui Chen}{0009-0006-0657-2749}

\affil{1}{Beihang University}

\correspond{*}{Corresponding author's email: 13682632518@buaa.edu.cn}

\newcommand{\NLanguagePairs}{38}
\newcommand{\NPrimaryTreebanks}{76}
\newcommand{\NCandidateTreebanks}{164}
\newcommand{\RawCCC}{0.391}
\newcommand{\RawCCCLow}{-0.096}
\newcommand{\RawCCCHigh}{0.715}
\newcommand{\RawICC}{0.399}
\newcommand{\RatioCCC}{0.508}
\newcommand{\RatioCCCLow}{0.220}
\newcommand{\RatioCCCHigh}{0.716}
\newcommand{\RatioICC}{0.515}
\newcommand{\RawMAE}{0.194}
\newcommand{\RatioMAE}{0.075}
\newcommand{\HOneCCCDifference}{0.117}
\newcommand{\HOneCCCDifferenceLow}{-0.240}
\newcommand{\HOneCCCDifferenceHigh}{0.574}
\newcommand{\HTwoRawContrast}{0.166}
\newcommand{\HTwoRawContrastLow}{0.114}
\newcommand{\HTwoRawContrastHigh}{0.249}
\newcommand{\HTwoRatioContrast}{0.067}
\newcommand{\HTwoRatioContrastLow}{0.047}
\newcommand{\HTwoRatioContrastHigh}{0.094}
\newcommand{\HThreeRawReduction}{0.033}
\newcommand{\HThreeRawReductionLow}{-0.004}
\newcommand{\HThreeRawReductionHigh}{0.085}
\newcommand{\HThreeRatioReduction}{0.040}
\newcommand{\HThreeRatioReductionLow}{0.022}
\newcommand{\HThreeRatioReductionHigh}{0.066}
\newcommand{\FamilyRawCCC}{-0.129}
\newcommand{\FamilyRatioCCC}{0.664}
\newcommand{\AllPairsRawCCC}{0.478}
\newcommand{\AllPairsRatioCCC}{0.415}
\newcommand{\NAllPairRows}{277}

\begin{document}
\maketitle

%==================================================
% Abstract

\begin{abstract}
Dependency-distance estimates derived from a single corpus are routinely treated
as properties of a language, yet this assumption has not been tested across
independently compiled corpora.
We compared mean dependency-distance estimates across \NLanguagePairs{}
same-language treebank pairs from Universal Dependencies~v2.18, using
concordance correlation, Bland--Altman analysis, and a twelve-specification
multiverse design. Cross-treebank agreement was moderate at best: substituting
one treebank for another reversed nearly 40\% of pairwise language orderings,
and treebank choice accounted for roughly 29\% of between-group variance.
This disagreement substantially exceeded within-treebank sampling error
and persisted across all twelve preprocessing specifications. Nevertheless,
every treebank confirmed dependency-length minimization (normalized ratio
below~1). The data are more consistent with MDD as a corpus-conditioned
composite of grammatical, register, and annotation factors than as a stable
language-level parameter: the qualitative DLM universal survives corpus
substitution, but the ordinal cross-linguistic ranking does not.
\end{abstract}

%==================================================
% Keywords

\begin{keywords}
dependency distance, measurement reliability, Universal Dependencies, multiverse analysis, reproducibility
\end{keywords}

%==================================================
% Introduction

\section{Introduction}
\label{sec:introduction}

Syntactic relations in natural language must be resolved across linear distance.
A comprehender encountering a syntactic head must retrieve its dependents from
working memory, and the cost of that integration grows with the number of
intervening elements \parencite{gibson1998linguistic,gibson2026dependency}. Producers face a complementary pressure:
grammars and production choices that keep syntactically related words close
together reduce processing load for both speaker and listener
\parencite{hawkins2004efficiency,gibson2019efficiency}. This trade-off between communicative demands and
processing constraints has made dependency distance, the linear separation
between a syntactic head and its dependent, a central quantity in quantitative
and computational linguistics. Mean dependency distance (MDD), averaged over all dependencies in a corpus,
provides a simple, cross-linguistically applicable measure of this processing
cost \parencite{liu2017dependency}. Observed dependency distances are consistently shorter than random-ordering
baselines, a pattern first demonstrated for individual languages
\parencite{ferrer2004euclidean,gildea2010grammars} and later confirmed across 37
typologically diverse languages using Universal Dependencies (UD) treebanks
\parencite{futrell2015largescale,futrell2020locality}, establishing dependency-length minimization (DLM)
as a candidate cross-linguistic universal \parencite{temperley2018minimizing}, though
\textcite{yadav2022reappraisal} argued that the DLM signal confounds word-order
preferences with other structural pressures such as heavy-constituent shift,
raising the possibility that MDD conflates multiple sources at the level of
individual sentences as well as at the level of corpora.

Because MDD depends on sentence length \parencite{ferrer2004euclidean} and is
further shaped by genre composition and annotation conventions, its value for
cross-linguistic comparison rests on whether the estimate generalizes beyond the
corpus that produced it. Yet applied studies routinely treat a single-corpus
estimate as representative. \textcite{chen2022texttypes} compared dependency
distance across text types within a single treebank; \textcite{li2026distance}
ranked languages by syntactic distance using one UD treebank per language; and
similar single-corpus designs underpin L2 proficiency assessment
\parencite{hao2022probability}, genre comparison
\parencite{chen2024czech,wang2023genre}, word-order typology
\parencite{liu2010dependency}, and recent extensions to hierarchical distance
\parencite{wang2025tradeoff} and communicative-efficiency normalization
\parencite{lei2026alpha}. This practice extends to the foundational DLM work
itself: \textcite{futrell2015largescale} used one treebank per language to
establish DLM universality. Universal Dependencies now offers multiple
independently compiled treebanks for dozens of languages
\parencite{demarneffe2021universal}, but whether MDD estimates agree across
different corpora has not been directly tested---a gap that recent calls for
reproducibility in corpus linguistics make increasingly pressing
\parencite{schweinberger2025reproducibility,flanagan2025reproducibility}.

What is at stake goes beyond measurement convenience. If MDD
primarily reflects stable grammatical constraints, corpus choice adds only
minor sampling noise, and treebanks from different sources should yield
concordant estimates. If, however, MDD is a composite---grammatical ordering
preferences set a structural baseline, register determines which structures are
sampled, and annotation conventions determine how those structures are
represented---then separately compiled corpora should agree only to the extent
that they sample the same mixture of components
\parencite{yadav2022reappraisal}. The degree of cross-treebank agreement
discriminates between these predictions.

The closest methodological precedent is \textcite{berdicevskis2018ud}, who split
individual UD treebanks into random halves and tested whether complexity
measures differed between subsamples. Split-half subsamples, however, share
source selection, genre composition, and annotation history, so their stability
does not establish agreement across distinct corpora
\parencite[cf.][]{biber2024organization}---and MDD was not their focal measure.

The study addresses three research questions:

\begin{enumerate}
  \item To what extent do raw and normalized dependency-distance estimates agree
  across distinct-source-group UD treebanks of the same language?
  \item How large is cross-treebank disagreement relative to finite-sample
  variation within each treebank, and how much is reduced by matching sentence
  lengths?
  \item How sensitive are agreement estimates to declared choices about
  punctuation, sentence-length limits, and aggregation?
\end{enumerate}

We test the directional hypotheses that random-order normalization increases
absolute agreement (H1), that cross-treebank disagreement exceeds variation in
500-sentence subsamples (H2), and that matching sentence-length distributions
reduces absolute disagreement (H3). These hypotheses were frozen before the full
metric run but were not externally registered; null or contrary results do not
trigger a change in the estimand or analysis set.

The present study treats same-language UD treebanks assigned to distinct
automated source groups as repeated measurements and evaluates their agreement
using concordance correlation, Bland--Altman analysis, and a
specification-curve design
\parencite{simonsohn2020specification,steegen2016multiverse}.
Cross-treebank agreement is moderate at best: disagreement substantially
exceeds within-treebank sampling error and reverses nearly 40\% of pairwise
language rankings. The data favor MDD as a corpus-conditioned composite whose
qualitative properties survive corpus substitution but whose ordinal
cross-linguistic ranking does not. All data, code, and numerical results are
generated by one reproducible pipeline.

%==================================================
% Data

\section{Data}
\label{sec:data}

Evaluating cross-treebank agreement requires, first, that each language be
represented by at least two treebanks and, second, that the two treebanks come
from genuinely different textual sources. This section describes the data
sources, the sentence-level eligibility criteria, and the automated procedure
used to separate shared-source treebanks from independent ones.

The study uses Universal Dependencies version 2.18, released on 15 May 2026
(handle: \url{http://hdl.handle.net/11234/1-6149}), and Glottolog CLDF 5.3 for
language-family metadata. Treebank data are retrieved from the checksum-verified
official release archive. The live UD index supplies metadata, but only
directories present in the frozen archive can enter the study. Every data source
is recorded with its URL, version, license, size, MD5 digest, and SHA-256
checksum.

\subsection{Inclusion criteria}
\label{sec:inclusion}

The metadata screen begins with treebanks reporting at least 500 sentences and
language codes represented by at least two such treebanks. From
\NCandidateTreebanks{} candidate treebanks meeting these initial counts, we apply
sentence-level eligibility filters: each sentence must contain 5--40
non-punctuation syntactic words under the canonical specification, a single
basic-dependency root, integer token identifiers, and a connected acyclic
dependency tree after punctuation removal. Multiword-token range rows and empty
nodes are not counted as syntactic words. Sentences containing up to 80
non-punctuation words are retained separately for the declared length-cap
sensitivity analysis. Every exclusion is counted by treebank and reason; the
complete flow appears in the supplementary material.

\subsection{Source-independence audit}
\label{sec:source-audit}

Because two treebanks of the same language may share source texts (through
common translation projects, shared web crawls, or derivative
corpora), independence is audited before primary pairs are selected. For every
pair of same-language treebanks, we calculate overlap in normalized sentence-text
hashes and in sentence identifiers. Two treebanks enter the same source
group if text-hash overlap reaches 5\% of the smaller corpus, or if sentence-ID
overlap reaches 80\% while their sentence-count ratio lies between 0.95 and 1.05.
Connected components of this overlap graph define source groups. The largest
eligible treebank in each group becomes its representative, and the two largest
distinct-group representatives form the primary pair for each language.

This procedure is an operational screen, not proof of full provenance
independence: corpora built from different source texts can still share register
composition, time period, or annotation guidelines. The 5\% threshold is an operational default; to assess its
sensitivity, we report agreement under thresholds from any exact
overlap through 20\%. Borderline cases (pairs with 1--10\% text
overlap) are reserved for manual inspection.

\subsection{Pair selection and metadata}
\label{sec:pair-selection}

After eligibility screening and source-overlap grouping, the procedure yielded
\NLanguagePairs{} language-level pairs comprising \NPrimaryTreebanks{} treebanks
from distinct source groups. Each pair's genre labels, eligible
sentence counts, license information, and source-group assignment are reported in
\autoref{tab:primary-agreement} and the supplementary pair table. The paired
treebanks span 10 language families, with Indo-European dominating the eligible
sample (24 of 38 languages). This imbalance reflects the current UD coverage and
limits the genealogical diversity of any generalization. With the analysis sample
defined, we next specify how dependency distance is measured and how agreement
between the two treebanks in each pair is quantified.

%==================================================
% Measures and Analysis

\section{Measures and Analysis}
\label{sec:methods}

This section defines the estimands, agreement statistics, and pre-specified
analyses.

\subsection{Dependency-distance metrics}
\label{sec:metrics}

For a retained dependency arc from head position $h_i$ to dependent position
$d_i$, raw sentence-level MDD is

\begin{equation}
  \mathrm{MDD} = \frac{1}{m}\sum_{i=1}^{m} |h_i-d_i|,
  \label{eq:mdd}
\end{equation}

where $m$ is the number of non-root arcs retained after preprocessing.
This quantity is the most widely reported dependency-distance summary in
quantitative linguistics
\parencite{hao2022probability,chen2022texttypes,chen2024czech}.

A raw MDD value, however, depends on sentence length: longer sentences allow
longer dependencies even under random word order. To separate the structural
signal from this length baseline, we normalize by the exact expected distance
under a uniformly random linear ordering of $n$ words, $(n+1)/3$
\parencite{ferrer2004euclidean}. The normalized
measure is

\begin{equation}
  R_{\mathrm{DD}} = \frac{\mathrm{MDD}}{(n+1)/3}.
  \label{eq:random-ratio}
\end{equation}

This analytic expectation avoids Monte Carlo error. For aggregation, the canonical treebank
estimate gives each eligible sentence equal weight. The token-weighted
specification, included in the analysis grid, weights each sentence's
contribution by the number of analyzed arcs.

\subsection{Agreement framework}
\label{sec:agreement}

The two treebanks in each pair are treated as exchangeable repeated measurements
of the same language under different corpus conditions, with neither serving as
a reference standard. No single statistic captures all dimensions of agreement,
so we report four complementary quantities, each addressing a distinct aspect:
concordance correlation measures absolute agreement on the original scale,
intraclass correlation adjusts for range restriction, Bland--Altman limits
translate disagreement into practical substitutability bounds, and the
specification curve (\autoref{sec:specifications}) audits how agreement varies
across researcher-controlled preprocessing decisions.

\textit{Concordance correlation coefficient} (CCC; \cite{lin1989ccc})
combines rank association with a penalty for departure from the identity line:

\begin{equation}
  \rho_c = \frac{2\rho\,\sigma_1\sigma_2}
               {\sigma_1^2 + \sigma_2^2 + (\mu_1 - \mu_2)^2},
  \label{eq:ccc}
\end{equation}

where $\rho$ is the Pearson correlation between the two treebank
estimates, $\sigma_1$ and $\sigma_2$ are their standard deviations, and $\mu_1$
and $\mu_2$ are their means. The denominator penalizes both scale shift
($\sigma_1 \neq \sigma_2$) and location shift ($\mu_1 \neq \mu_2$), so
$\rho_c = 1$ only when the paired values fall on the identity line.

\textit{Intraclass correlation} ICC(1,1) \parencite{shrout1979icc}, the one-way
random single-measure form, is

\begin{equation}
  \text{ICC}(1,1) = \frac{\text{MS}_B - \text{MS}_W}
                         {\text{MS}_B + (k-1)\,\text{MS}_W},
  \label{eq:icc}
\end{equation}

where $\text{MS}_B$ and $\text{MS}_W$ are the between-language and
within-language mean squares from a one-way ANOVA and $k = 2$ is the number of
treebanks per language. This form is used because each language has a different
pair of treebanks; ICC forms that assume the same named instruments across all
languages are inapplicable.

\textit{Bland--Altman analysis} \parencite{bland1986agreement} plots the difference
between paired estimates against their mean. The 95\% limits of agreement,

\begin{equation}
  \text{LoA} = \bar{d} \pm 1.96\,s_d,
  \label{eq:loa}
\end{equation}

where $\bar{d}$ is the mean pairwise difference and $s_d$ its standard
deviation, define the range within which the two treebank estimates are expected
to fall. These limits assume approximate normality of pairwise differences and
serve as descriptive summaries; the BCa bootstrap provides the primary
uncertainty quantification. \textit{Spearman rank correlation} is retained as a description of rank
stability but is not treated as a measure of agreement.

Standard reliability benchmarks \parencite[moderate: 0.50--0.75; good: 0.75--0.90;][]{cicchetti1994guidelines} are reported for orientation but were developed
for controlled inter-rater contexts and may not transfer to corpus estimates; we
do not adopt a fixed acceptability threshold. Primary uncertainty intervals use 20{,}000 BCa bootstrap samples of language
pairs; the analysis grid uses 5{,}000 per specification. The raw-versus-normalized
contrast (H1) uses the same resampled pairs to ensure a paired comparison.
Family-cluster bootstrap intervals and equal-family-weight estimates assess
whether genealogical dependence alters the conclusions.

\subsection{Sampling stability and sentence-length matching}
\label{sec:sampling-matching}

Before attributing cross-treebank disagreement to corpus-construction
differences, we must rule out a simpler explanation: that the disagreement
is merely sampling noise amplified by small treebanks. Sampling stability
quantifies how much a treebank estimate fluctuates when only a subset of its
sentences is used. For every primary treebank, we draw 50, 100, 250, and 500
sentences without replacement, repeat each draw 500 times, and record the bias
relative to the full estimate, standard deviation, coefficient of variation, and
percentile intervals. The subsample sizes span the range from the smallest
eligible treebanks in the study ($\sim$500 sentences) to an order of magnitude
below the largest, ensuring that the stability curve covers the conditions
actually present in the data. Where at least 80\% of eligible sentences carry
document identifiers and at least five documents are available, complete
documents are sampled rather than individual sentences, because sentences within
a document share topic, register, and syntactic priming effects that
sentence-level sampling would understate; results from treebanks without
sufficient document metadata are explicitly marked as sentence-level fallbacks.

A second confound is sentence-length composition. Because raw MDD is
mechanically sensitive to sentence length (longer sentences permit longer
dependencies even under random word order), two treebanks that differ in their
sentence-length distributions will disagree in MDD partly for this arithmetic
reason alone. To isolate this component, we test whether aligning the distributions reduces
cross-treebank disagreement. The H3 estimand is

\begin{equation}
  \Delta_{\text{H3}} = \text{MAD}_{\text{unmatched}}
    - \overline{\text{MAD}}_{\text{matched}},
  \label{eq:h3}
\end{equation}

where $\text{MAD}_{\text{unmatched}} = |m_B - m_A|$ is the absolute
disagreement between full-treebank means and
$\overline{\text{MAD}}_{\text{matched}}$ is the mean absolute disagreement across
1{,}000 length-matched draws. For each draw, we sample 50 sentences from each of
four length bins (5--7, 8--10, 11--15, and 16--20 non-punctuation words) from
both treebanks. The bin widths increase with sentence length (3, 3, 5, and 5
words) to maintain roughly balanced sentence counts across bins: under the
canonical specification, the four bins contained 24\%, 22\%, 32\%, and 23\% of
sentences in the 5--20 word range, ensuring adequate coverage for the
within-bin sampling in each draw. The cap at 20 words limits the comparison to
the sentence-length range where most pairs have sufficient sentences per bin,
though some pairs with skewed length distributions may still lack adequate
coverage in one or more bins. A
language with an insufficient bin remains in the primary agreement analysis but
is transparently excluded from this matched comparison.

\subsection{Specification analysis}
\label{sec:specifications}

Analytic choices in preprocessing are not innocuous. To make their influence
transparent, we report agreement under every combination in the pre-declared
analysis grid: punctuation included or excluded, maximum non-punctuation sentence
length of 20, 40, or 80 words, and sentence-weighted or token-weighted
aggregation. The Cartesian product yields twelve specifications. We report the
full distribution of agreement estimates across these specifications and do not
apply an arbitrary percentage threshold to relabel a continuous pattern as
``robust'' or ``fragile'' \parencite{simonsohn2020specification,steegen2016multiverse}. The canonical
specification (punctuation excluded, 40-word cap, sentence-weighted) was
declared before the full metric run.

\subsection{Diagnostic and sensitivity analyses}
\label{sec:diagnostics-methods}

Beyond the primary agreement estimates and the specification grid, several
pre-declared diagnostic analyses test whether the conclusions depend on specific
design choices.

\paragraph{Sampling-error contrast.}
For each language pair, the H2 estimand is

\begin{equation}
  \Delta_{\text{H2}} = |m_B - m_A|
    - \hat{E}\bigl[|\epsilon_A - \epsilon_B|\bigr],
  \label{eq:h2}
\end{equation}

where $m_A$ and $m_B$ are the full-treebank means and each
$\epsilon = \bar{x}_{\text{sub}} - m$ is the deviation of a 500-sentence
subsample mean from its own full-treebank mean $m$. The expectation is estimated
by simultaneously drawing 500 sentences without replacement from each treebank
(chosen to fall below the smallest eligible treebank, 774 sentences, so that
the baseline reflects worst-case finite-sample conditions in the study),
computing $|\epsilon_A - \epsilon_B|$, and averaging over 500 repetitions.
$\Delta_{\text{H2}} > 0$ indicates that full-treebank disagreement exceeds what
finite sampling at 500 sentences would predict.

\paragraph{Sample composition.}
To assess whether the Indo-European dominance of the eligible sample drives the
aggregate, we recompute CCC under equal total weight per language family
(Glottolog top-level) and report family-cluster BCa bootstrap intervals using
the family as the resampling unit. As a supplement to the primary
two-per-language design, we also form every eligible within-language treebank
pair; because pair rows within a language share treebanks, each language receives
equal total weight and the language remains the resampling unit.

\paragraph{Metadata and comparability checks.}
Genre-label Jaccard overlap between paired treebanks is regressed on absolute
disagreement via OLS with HC3 standard errors and family-stratified permutation
testing. Treebank-substitution sensitivity replaces Treebank A with the
second-largest treebank for each language and records the proportion of pairwise
language orderings that reverse. Source-overlap threshold sensitivity reports
agreement under thresholds from any exact text match through 20\%, and a
separate analysis removes every exactly shared sentence hash and recomputes
agreement on the remaining data. All primary pairs are additionally reviewed for
strict comparability on variety, time period, speaker population, translation
status, and genre; pairs meeting all criteria are reported separately as an
illustrative check on disagreement when metadata dimensions are approximately
matched.

%==================================================
% Results

\section{Results}
\label{sec:results}

\subsection{Analysis sample}
\label{sec:results-sample}

After sentence-level eligibility screening and source-overlap grouping, the
primary analysis contained \NLanguagePairs{} languages and \NPrimaryTreebanks{}
treebanks drawn from distinct source groups. Source grouping combined
the 5\% text-overlap threshold with manual verification of five borderline
pairs whose text overlap fell between 1\% and 10\%: three were confirmed as
sharing a textual source and reclassified into the same group; two were confirmed
as independent. A random sample of 30 sentences was manually re-computed for MDD;
all 30 agreed exactly with the pipeline output.

The 38 primary pairs span 10 Glottolog families, but the distribution is heavily
skewed: 24 languages belong to Indo-European, four to Afro-Asiatic, and two each
to Uralic and Turkic, with the remaining four families represented by a single
language each. Treebank sizes range from 774 to over 175{,}000 eligible sentences
under the canonical specification. Genre labels, reported from UD metadata, vary
widely across pairs: some pairs share their primary genre (e.g., news for
Arabic), while others differ markedly (e.g., spoken versus blog-and-web for
English).

Manual comparability review of all \NLanguagePairs{} pairs assessed whether each
pair matched on variety, time period, speaker population, translation status, and
genre. Only three pairs (Arabic, Armenian, and Italian) met all criteria; the
remaining 35 differ in at least one dimension. Perfect comparability is the
exception, not the norm. This heterogeneity is characteristic of UD's multi-treebank coverage. The
agreement estimates that follow should therefore be read as upper bounds on
reproducibility under convenience-corpus substitution, not as estimates of
agreement between carefully matched samples.

\subsection{Cross-treebank agreement}
\label{sec:results-agreement}

The central question is whether two independently sourced treebanks of the same
language produce similar dependency-distance estimates. If corpus choice were a
minor nuisance, we would expect CCC near 1 and negligible mean differences. The
data show neither. Under
the canonical specification, raw MDD showed CCC = \RawCCC{} (95\% BCa interval:
[\RawCCCLow{}, \RawCCCHigh{}]) and ICC(1,1) = \RawICC{}; the normalized ratio
showed CCC = \RatioCCC{} ([\RatioCCCLow{}, \RatioCCCHigh{}]) and ICC(1,1) =
\RatioICC{}. Two distinct-source treebanks of the same language differed by a
mean of \RawMAE{} words in raw MDD and \RatioMAE{} in the normalized
ratio, roughly a 9\% shift relative to the grand mean for raw MDD. We report
ICC(1,1) alongside CCC because CCC can be depressed by range restriction when
the sample spans a narrow band of true values; the similar ICC values confirm
that the moderate agreement is not an artifact of limited cross-language
variance but reflects genuine between-treebank discrepancy.

\begin{figure}[t]
\centering
\includegraphics[width=\textwidth]{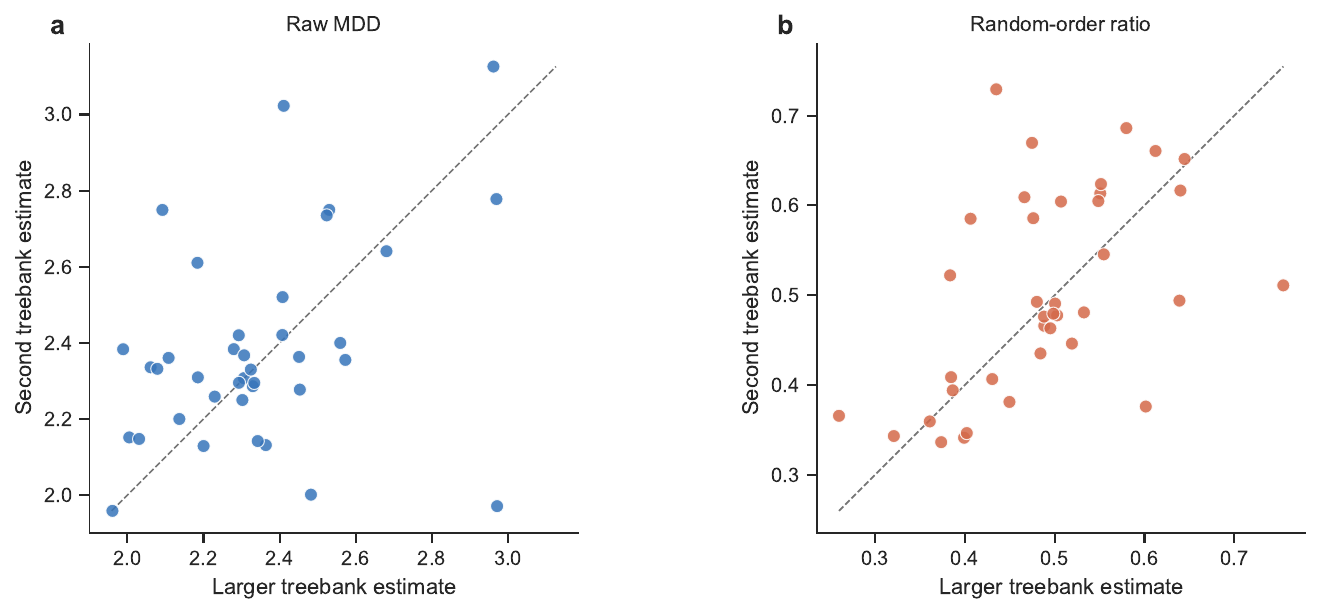}
\caption{Paired dependency-distance estimates for each language under the
canonical specification. Each point represents one language; the dashed line
marks exact agreement.}
\label{fig:paired-estimates}
\end{figure}

\autoref{fig:paired-estimates} plots the paired treebank estimates against the
identity line. Several languages cluster near the diagonal, but Irish (ga),
Chinese (zh), Polish (pl), and English (en) show large departures. Irish, for
instance, pairs a mixed-genre national-language treebank with a Twitter corpus, a
register mismatch that plausibly accounts for much of the observed disagreement.

\begin{table}[ht]
\center
\caption{Cross-treebank agreement under the canonical specification (punctuation
excluded, 40-word cap, sentence-weighted). CCC = concordance correlation
coefficient; ICC = intraclass correlation, one-way random single measure;
MAE = mean absolute difference; LoA = 95\% limits of agreement.}
\label{tab:primary-agreement}
\begin{tabular}{lrcrrrr}
\toprule
\textbf{Measure} & \textbf{CCC} & \textbf{95\% CI} & \textbf{ICC(1,1)} & \textbf{MAE} & \textbf{Bias} & \textbf{LoA} \\
\midrule
Raw MDD & 0.391 & [$-$0.096, 0.715] & 0.399 & 0.194 & 0.036 & [$-$0.522, 0.595] \\
Normalized ratio & 0.508 & [0.220, 0.716] & 0.515 & 0.075 & 0.013 & [$-$0.193, 0.219] \\
\bottomrule
\end{tabular}
\end{table}

The paired H1 contrast (the difference in CCC between the normalized ratio and
raw MDD, computed on the same bootstrap draws) was \HOneCCCDifference{} (95\%
BCa interval: [\HOneCCCDifferenceLow{}, \HOneCCCDifferenceHigh{}]). Because the
interval spans zero and extends from a substantial negative to a substantial
positive value, these data did not establish that random-order normalization
improved absolute agreement under the canonical specification. The point estimate favors normalization, but the uncertainty is wide enough
that the reverse conclusion is also compatible with the data.

\begin{figure}[t]
\centering
\includegraphics[width=\textwidth]{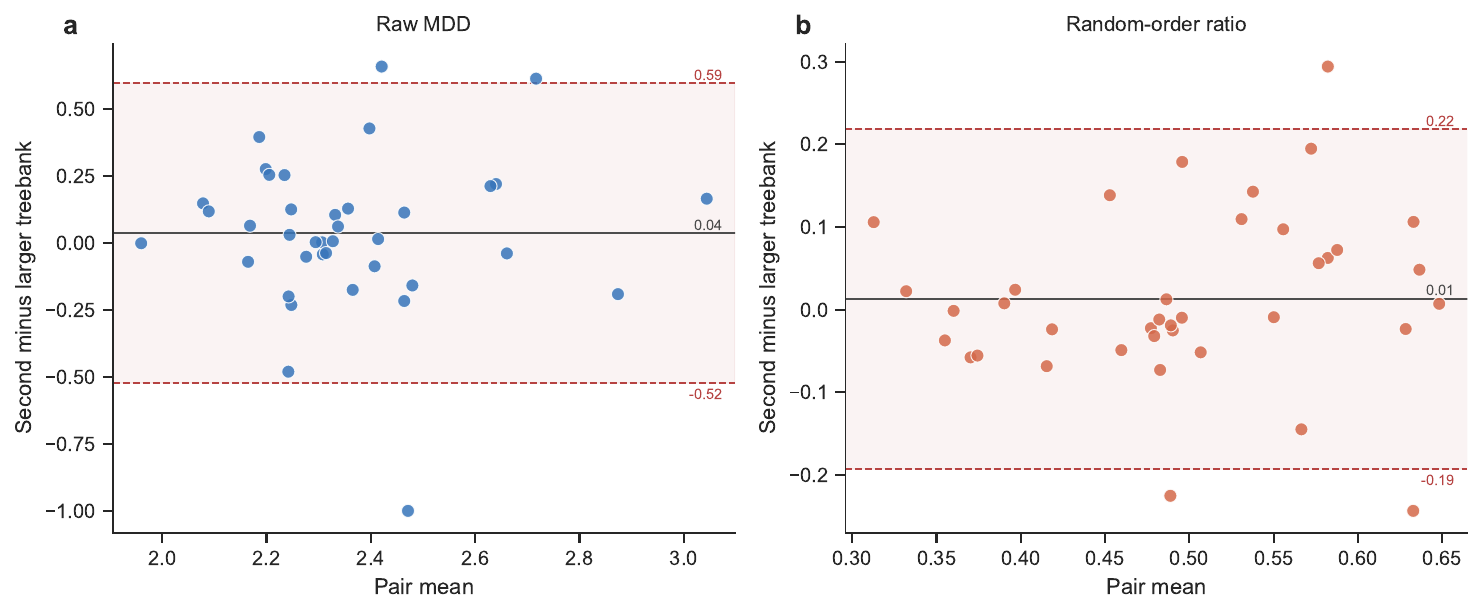}
\caption{Bland--Altman assessment of cross-treebank agreement. The horizontal
axis shows the mean of each pair's two treebank estimates; the vertical axis
shows their difference (second treebank minus larger treebank). Solid lines mark mean
bias; dashed lines mark 95\% limits of agreement.}
\label{fig:bland-altman}
\end{figure}

\autoref{fig:bland-altman} presents the Bland--Altman diagnostic. For raw MDD, the
mean bias was 0.036 (the second-largest treebank tended to yield slightly higher estimates),
and the 95\% limits of agreement spanned [$-$0.522, 0.595], a range of over one
word of dependency distance. In the observed data, a language whose MDD was 2.3 in one treebank could
yield a value anywhere between roughly 1.8 and 2.9 in its independently
sourced partner. The normalized ratio showed narrower limits
([$-$0.193, 0.219]) and near-zero bias (0.013), but even here the 95\% range
covers about 0.41 units, a non-trivial interval given that most language-level
estimates cluster between 0.3 and 0.7. In neither case does the scatter suggest
strong heteroscedasticity; disagreement does not systematically increase with
the magnitude of the estimate. One practical consequence of this moderate agreement is quantified in
\autoref{sec:results-diagnostics}: substituting one treebank for another reverses
nearly 40\% of pairwise language orderings.

Despite this moderate concordance, the qualitative conclusion of
dependency-length minimization is robust: the normalized ratio was below~1.0
for all 76 primary treebanks (range: 0.26--0.76), confirming that observed
dependency distances are shorter than random-ordering baselines in every
corpus. Cross-treebank disagreement affects where a language falls in the
ranking, not whether it minimizes dependency length.

\subsection{Decomposing disagreement: sampling error and sentence length}
\label{sec:results-sampling}

The H2 test asks whether the cross-treebank disagreement observed for each
language exceeds the variation expected from drawing only 500 sentences from each
treebank. Across the 38 language pairs, observed absolute disagreement exceeded
the 500-sentence joint sampling error by \HTwoRawContrast{} for raw MDD (95\%
BCa interval: [\HTwoRawContrastLow{}, \HTwoRawContrastHigh{}]) and by
\HTwoRatioContrast{} for the normalized ratio ([\HTwoRatioContrastLow{},
\HTwoRatioContrastHigh{}]). Both intervals exclude zero: finite-sample noise
at 500 sentences does not account for the between-treebank differences.

Collecting more sentences from the same source would not resolve the disagreement
between independent treebanks; the data point to source-level
differences rather than finite-sample noise as the dominant contributor.

Sentence-length matching was feasible for 30 of 38 language pairs (the remaining
eight had insufficient coverage in at least one length bin). For these 30 pairs,
matching reduced absolute disagreement in the normalized ratio by
\HThreeRatioReduction{} ([\HThreeRatioReductionLow{},
\HThreeRatioReductionHigh{}]), supporting H3. Relative to the baseline ratio MAE
of \RatioMAE{}, this reduction represents roughly half of the total
disagreement, a practically meaningful effect. The corresponding reduction for raw
MDD was \HThreeRawReduction{} ([\HThreeRawReductionLow{},
\HThreeRawReductionHigh{}]), whose interval included zero. Length-distribution
differences thus explain a detectable portion of inter-treebank disagreement for
the normalized ratio, but the effect for raw MDD was not clearly established.

A nested variance decomposition (family/language/treebank as random intercepts,
with a cubic spline on sentence length; full models in Supplementary~S5)
quantifies how total variance partitions across levels. For raw MDD, the
variance components were: family 0.006, language 0.036, treebank 0.017, and
residual (sentence-level) 0.253. Treebank-level variance thus accounts for
roughly 29\% of the between-group variance (family + language + treebank),
confirming that corpus choice contributes a non-negligible share of the
systematic signal. For the normalized ratio, treebank-level variance was
effectively zero after the sentence-length adjustment, consistent with the H3
finding that sentence-length composition is a primary driver of inter-treebank
disagreement once the length baseline is removed.

Taken together, these results indicate that cross-treebank disagreement
cannot be attributed to finite sampling or sentence-length composition alone.
The residual reflects differences in register, source population, annotation
practice, or their interaction.

\subsection{Genealogical and sampling-frame diagnostics}
\label{sec:results-diagnostics}

The primary agreement estimates rest on specific design choices: the weighting
of languages, the overlap threshold, and which treebank represents each
language.

\paragraph{Sample composition.}
Because 24 of 38 languages belong to Indo-European, the aggregate may
primarily reflect one family's behavior. Under equal-family weighting
(10~clusters), CCC shifted to \FamilyRawCCC{} for raw MDD and
\FamilyRatioCCC{} for the normalized ratio, but with only ten independent
clusters these estimates are unstable: a single divergent family can dominate
the reweighted mean, and BCa coverage is unreliable. The shift is noted as
a sensitivity check, not a substantive finding.

The supplementary all-pairs analysis, which forms every eligible within-language
pair rather than selecting only the two largest source-group representatives,
contains \NAllPairRows{} pair rows distributed across \NLanguagePairs{} language
clusters. With equal total weight per language and language-cluster resampling,
CCC was \AllPairsRawCCC{} for raw MDD and \AllPairsRatioCCC{} for the
normalized ratio. These values do not represent independent observations: pair
rows share treebanks, and the resampling unit remains the language.

\paragraph{Metadata and comparability checks.}
Genre-label Jaccard overlap, an exploratory corpus-level descriptor, showed
adjusted OLS coefficients of $-$0.146 for raw MDD absolute disagreement and
$-$0.039 for the normalized ratio. Neither was statistically distinguishable
from zero under HC3 standard errors or family-stratified permutation testing.
A Jaccard score records shared UD genre labels rather than genre proportions
and cannot identify a causal register effect; it is reported here as a
covariate description, not a mechanistic explanation.

Treebank-substitution sensitivity, replacing Treebank A with the second-largest
treebank for each language and recalculating language rankings, reversed 39.5\%
of pairwise language orderings for raw MDD and 28.7\% for the normalized ratio.
The reversal rate and the moderate CCC capture different facets of the same
disagreement: CCC summarizes average absolute deviation from the identity line,
while ranking reversal records how often that deviation is large enough, relative
to the between-language spread, to swap ordinal positions. A CCC of 0.39 is
compatible with a 40\% reversal rate precisely because between-language variance
in raw MDD is modest; even small absolute shifts can reorder languages whose
estimates are closely spaced. Concretely, the mean inter-treebank shift
(\RawMAE{} words) is comparable to the between-language interquartile range
($\approx$0.28 words), so reversals concentrate among closely spaced languages
in the dense middle of the distribution rather than among outliers.
This means that for roughly four in ten language pairs, the relative ranking on
raw MDD would change if a different corpus were used. Cross-linguistic
comparisons of dependency distance, whether ranking languages by complexity
\parencite{berdicevskis2018ud,li2026distance} or testing typological
hypotheses, implicitly assume that the chosen corpus is representative enough
to fix a language's position in the ranking. A 40\% reversal rate calls that
assumption into question.

Tightening the overlap threshold to any exact text match (removing five additional
pairs) left 33 language pairs with CCC values of 0.353 (raw MDD) and 0.434
(normalized ratio). Removing every exactly shared sentence hash from the
remaining 38 pairs produced essentially unchanged agreement (CCC = 0.391 and
0.507); the small number of shared sentences detected did not
drive the primary estimates.

In summary, the primary agreement estimates are robust to the overlap threshold
and to residual shared sentences, but they are sensitive to genealogical
weighting and to which treebank represents each language.

\subsection{Sensitivity to preprocessing specifications}
\label{sec:results-specifications}

The preceding diagnostics held the preprocessing pipeline fixed and varied
the sample composition. A different source of researcher-controlled variation
is the preprocessing pipeline itself: which tokens count, how long a sentence
may be, and how sentence contributions are weighted. If agreement is fragile
to these decisions, the canonical estimate is one point in a distribution
rather than a stable summary.

\begin{figure}[t]
\centering
\includegraphics[width=\textwidth]{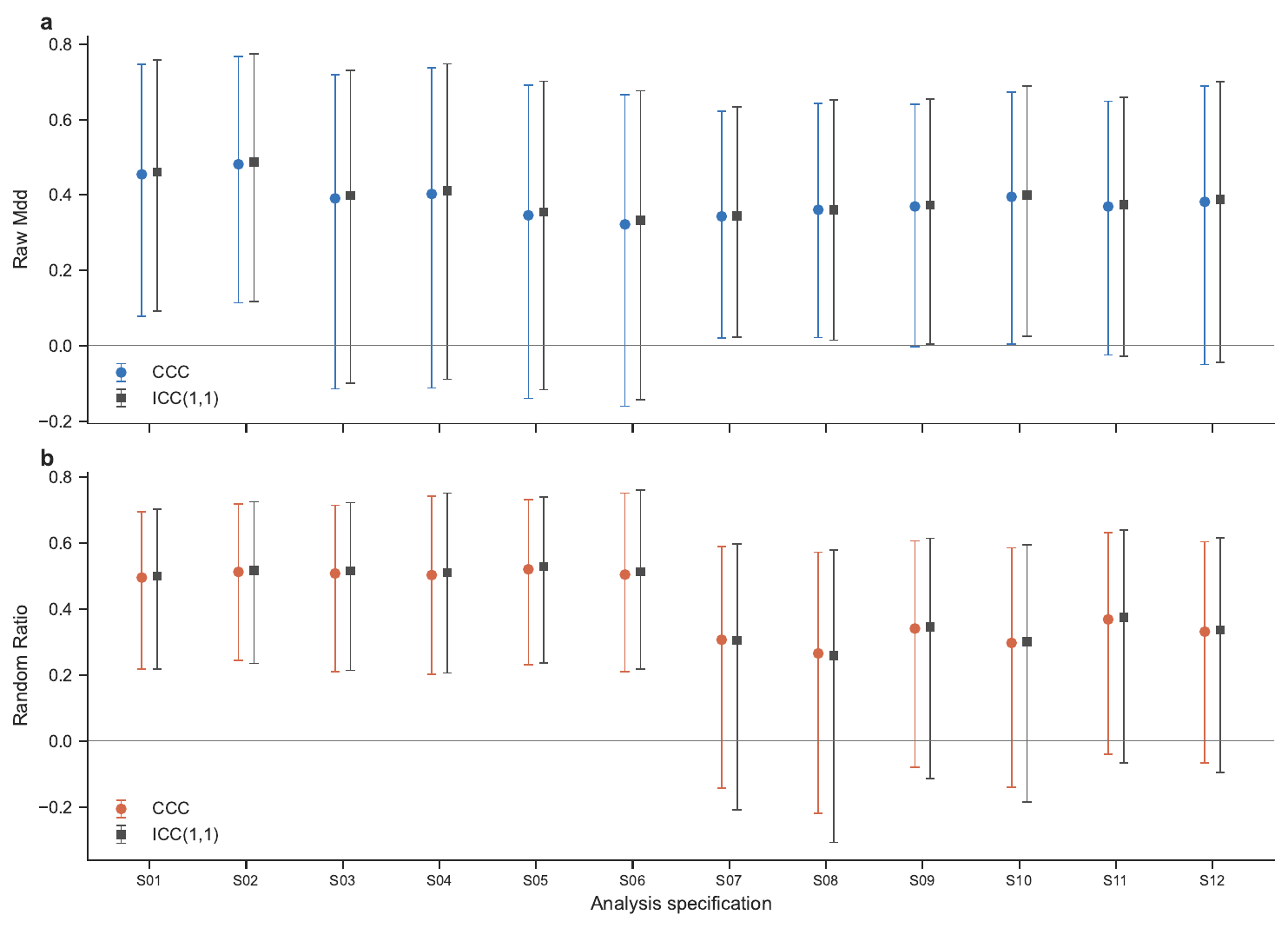}
\caption{Agreement estimates and bootstrap intervals across the twelve
pre-declared analysis-grid specifications. Each point is a CCC estimate;
vertical bars show 95\% BCa intervals. Specifications are ordered by CCC within
each metric.}
\label{fig:specification-curve}
\end{figure}

\autoref{fig:specification-curve} displays agreement estimates and their bootstrap
intervals across all twelve pre-declared specifications. Across the grid,
CCC for raw MDD ranged from 0.322 (punctuation excluded, 80-word cap,
token-weighted) to 0.481 (punctuation excluded, 20-word cap,
token-weighted). For the normalized ratio, CCC ranged from 0.265 (punctuation
included, 20-word cap, token-weighted) to 0.520 (punctuation excluded, 80-word
cap, sentence-weighted).

Two patterns are visible. First, for the normalized ratio, excluding punctuation
produced higher agreement than including it in every combination of length cap
and weighting. For raw MDD, the same direction held at the 20-word and 40-word
caps, but at the 80-word cap the pattern reversed: including punctuation yielded
higher CCC than excluding it. The reversal at 80 words likely reflects the
interaction between punctuation tokenization inconsistencies across treebanks
and the changing proportion of short-distance \texttt{punct} arcs at longer
sentence lengths. Second, the effect of sentence-length cap and weighting was modest and
non-monotonic: no single specification dominated all others.

The canonical specification (punctuation excluded, 40-word cap,
sentence-weighted) falls near the center of both distributions. The spread
across specifications is appreciable, roughly 0.16 CCC units for raw MDD and
0.26 for the normalized ratio, indicating that preprocessing choices contribute
meaningfully to the reported agreement. No specification produced CCC values
above 0.52, and no principled threshold for ``acceptable'' agreement exists in
quantitative linguistics. Any single-specification study of dependency distance
implicitly commits to one point in this range, and readers cannot assess
cross-study comparability without knowing which point was chosen.

\subsection{Agreement among strictly comparable pairs}
\label{sec:results-strict}

Every analysis above operates on the full 38-pair sample, in which most pairs
differ in genre, register, or time period. The natural objection is that
disagreement may simply reflect these metadata mismatches rather than corpus
construction per se. The three pairs that passed full comparability
review (Arabic, Armenian, and Italian) offer an illustrative
check: they show what disagreement looks like when register and variety are held
approximately constant, though three convenience pairs cannot establish a formal
bound.

\begin{table}[ht]
\center
\caption{Estimates for the three strictly comparable pairs under the canonical
specification.}
\label{tab:strict-comparable}
\begin{tabular}{llrrr}
\toprule
\textbf{Language} & \textbf{Metric} & \textbf{Treebank A} & \textbf{Treebank B} & \textbf{Difference} \\
\midrule
Arabic & Raw MDD & 2.229 & 2.259 & $-$0.030 \\
Arabic & Normalized ratio & 0.321 & 0.343 & $-$0.022 \\
Armenian & Raw MDD & 2.293 & 2.295 & $-$0.002 \\
Armenian & Normalized ratio & 0.480 & 0.493 & $-$0.013 \\
Italian & Raw MDD & 2.185 & 2.310 & $-$0.124 \\
Italian & Normalized ratio & 0.430 & 0.407 & 0.024 \\
\bottomrule
\end{tabular}
\end{table}

Arabic and Armenian showed small absolute differences (0.030 and 0.002 in raw
MDD), within the range that sampling error alone might produce. Italian, despite
passing comparability screening, showed a larger raw-MDD difference of 0.124,
comparable to the median absolute disagreement across all 38 pairs (0.137).
\textcite{chen2022texttypes} demonstrated that dependency distance varies
substantially across text types even within a single language; the Italian
result suggests that residual genre or source differences below the threshold of
metadata labels can produce effects of similar magnitude.

Three pairs cannot anchor a reliable subgroup estimate, but the direction is
consistent: even when metadata suggest comparability, corpus construction
can leave a measurable imprint on the estimate.

%==================================================
% Discussion

\section{Discussion}
\label{sec:discussion}

The preceding results establish that cross-treebank disagreement is moderate
and exceeds within-treebank sampling error. On the stable-parameter view,
corpus choice should add only minor sampling noise and treebanks should yield
high concordance; the data do not support that prediction. We organize the
discussion around
four questions: what drives the disagreement beyond sampling error, whether
normalization helps, how sensitive results are to preprocessing choices, and
what the pattern implies for the nature of MDD as a construct and for
comparative research practice.

\subsection{Why cross-treebank disagreement exceeds sampling error}

The H2 test excluded finite sampling at 500 sentences as a sufficient
explanation, but it did not identify which source-level factors drive the
residual. Our design cannot decompose annotation practice, register, time period,
translation status, or cross-variety differences, but two candidates are
independently documented in the UD literature.

\paragraph{Annotation divergence.}
\textcite{zeldes2023consistent} examined the two largest English UD treebanks (EWT
and GUM) across six release versions and found systematic differences in
tokenization, tagging, and dependency analysis that persisted even as
cross-corpus parsing accuracy improved. If two treebanks of the same
language, both following the same guidelines, still diverge in how they handle
compounds, clausal subjects, and multiword tokens, it follows that aggregate
statistics built on those annotations, including MDD, could diverge as well.

\paragraph{Register and genre composition.}
\textcite{dobrovoljc2026counting} found that spoken and written English corpora
draw on largely non-overlapping inventories of syntactic structure types,
and within-language text-type variation in dependency distance is independently
documented \parencite{chen2022texttypes}. In our sample, most treebank pairs differ
in genre metadata (\autoref{sec:results-agreement}), and the data pattern points more
readily to register as the dominant driver than to annotation alone: the
largest departures from the identity line involve pairs with overt register
mismatches (Irish: national-language treebank vs.\ Twitter; Chinese: news vs.\
learner text), whereas the three comparable pairs showed small differences for
Arabic and Armenian. If annotation divergence were the sole source, the size of these
register-linked departures would be difficult to explain. To be sure,
nominally shared UD guidelines still permit substantial annotation
variability in practice \parencite{zeldes2023consistent}, so annotation and register
effects are confounded in our design rather than cleanly separated.
Our genre-label Jaccard predictor did not reach statistical significance
(\autoref{sec:results-diagnostics}), but this is compatible with the Jaccard
measure being too coarse rather than with absence of a register effect.
Whether register, annotation, or their interaction dominates, the pattern
supports MDD as a composite quantity whose value is conditioned on the corpus
that produced it.

\subsection{Why normalization does not clearly improve agreement}

Register and annotation, the two candidates identified above, are distinct
from the sentence-length composition that random-order normalization targets.
On the composite account, if these non-length sources dominate, normalization
alone should not resolve the disagreement --- and the wide H1 interval
(including both improvement and worsening) is consistent with precisely that
prediction.

The variance decomposition (\autoref{sec:results-agreement}) sharpens this
picture: after length adjustment, treebank-level variance in the normalized
ratio effectively vanished, whereas raw-MDD treebank variance remained
0.017. Normalization thus removes the sentence-length component of
corpus-conditioned variation but leaves register and annotation effects
intact. Two further observations qualify this picture. The sentence-length
matching analysis (H3) established that equalizing length distributions
reduced ratio disagreement by 0.040 (CI excluding zero), confirming that
sentence length is a detectable contributor. At the same time, under
equal-family weighting the ratio retained moderate agreement
(\FamilyRatioCCC) while raw MDD collapsed, suggesting that the length
correction absorbs some compositional differences between treebank pairs.
Taken together, these findings indicate that
normalization addresses one source of variation among several, placing a
ceiling on how much improvement any single-dimension correction can achieve.

\subsection{Preprocessing as a researcher degree of freedom}

The most consequential single preprocessing choice was punctuation treatment,
whose effect reversed direction at the 80-word cap for raw MDD. The
dependency-distance domain thus echoes a broader pattern:
\textcite{alves2023typology} demonstrated that four word-order typology
methods applied to 20 parallel-corpus languages produced different typological
clusterings from identical data. Our specification curve demonstrates the
dependency-distance analog: analytical decisions shift agreement estimates
by roughly 0.16 CCC units (raw) and 0.26 (ratio), a range comparable to the
difference between the two metrics themselves. On the composite view, this
sensitivity has a natural account: each preprocessing choice alters which
components of the signal enter the calculation. Including or excluding
punctuation changes the arc inventory and thus the weight of annotation
conventions; raising the sentence-length cap shifts the register composition of
the retained sample. Specification sensitivity is, in this light, a further
symptom of MDD's composite character.

\subsection{Implications for comparative research}

A key concern is whether corpus sensitivity undermines the cross-linguistic
evidence for dependency-length minimization (DLM). It does not: every one of
the 76 treebanks in our sample yielded a normalized ratio below~1
(\autoref{sec:results-agreement}), confirming that observed MDD falls below
random-order baselines regardless of which corpus represents a language
\parencite{futrell2015largescale,futrell2020locality}. The DLM universal survives
corpus substitution because it is an \emph{ordinal} fact about each language
individually, not a comparison across languages.

Cross-linguistic ordinal inference, by contrast, is less secure. Treebank-level variance
accounts for roughly 29\% of between-group variance in raw MDD
(\autoref{sec:results-agreement}), and corpus substitution reversed
nearly 40\% of pairwise rankings (\autoref{sec:results-diagnostics}). To put this in concrete terms, the mean inter-treebank
shift (\RawMAE{} words) is comparable to the between-language interquartile
range ($\approx$0.28 words), so reversals concentrate among closely spaced
languages rather than among outliers. These observations converge on the composite interpretation --- MDD as
shaped by grammar, register, and annotation in corpus-dependent
proportions --- but we stress that the interpretation rests on converging
circumstantial evidence, not a demonstrated decomposition; the design
confounds the three components and cannot quantify their individual
contributions.

Three practical recommendations follow. First, cross-linguistic comparisons
should use multiple treebanks per language where available and report the range
of estimates; where only one treebank exists, the Bland--Altman limits in
\autoref{sec:results-agreement} provide an empirical prior on corpus-conditioned
variation. Second, specification-curve analysis
\parencite{simonsohn2020specification,steegen2016multiverse} should accompany
treebank-derived conclusions, because analytical decisions shift agreement
estimates by roughly 0.16--0.26 CCC units
(\autoref{sec:results-specifications}). Third, researchers should report the
preprocessing chain (punctuation treatment, length cap, weighting) with the
same precision as the corpus name, so that readers can locate each study within
the specification space.

These findings raise a question that extends beyond MDD: does the sensitivity
we document reflect a general deficiency of treebank-derived measures, or
the composite character of this particular metric? The composite
interpretation generates a testable prediction: measures conflating more dimensions
should be more corpus-sensitive, while measures isolating a single dimension
should prove more stable. \textcite{moscoso2025entropy} found that derivational
entropy rate remains stable across centuries and registers within a single
annotation scheme, as expected for a measure isolating a distributional
property of word formation. \textcite{egbert2025stability} found that corpus-based
word-type frequency lists required substantially larger samples than typically
assumed, reinforcing that measurement instability extends beyond dependency
distance. These contrasts suggest that the reliability problem is not to
suppress noise around a fixed quantity but to separate the signal of interest
from the other signals that MDD conflates --- for instance by estimating
agreement within register-matched subsamples, or by designing measures that
factor out annotation-dependent arc types.

\subsection{Limitations}

Several design constraints qualify the conclusions above. With only ten
language-family clusters, the equal-family reweighting is unstable and its
results are treated as a sensitivity check rather than a finding. Our design
cannot decompose annotation practice, register, time period, and translation
status because these factors are confounded across treebank pairs; a direct
test would require a factorial design crossing annotation team with register
within a single language. More broadly, because the composite label is
motivated by converging patterns rather than by an identified decomposition,
it cannot be distinguished within the present design from a simpler account
in which low agreement reflects heterogeneity in UD treebank construction
rather than an inherent property of MDD as a measure. The genre-label Jaccard
measure is coarse and may understate true register distance, and the
source-independence audit (text-hash
overlap) captures verbatim duplication but not paraphrase or shared domain.
Only three language pairs qualified as strictly comparable, too few to anchor
an independent estimate. The 38-pair sample is dominated by Indo-European
languages (24 of 38); the composite interpretation should be tested against
a more genealogically balanced sample as UD coverage expands.

%==================================================
% Conclusion

\section{Conclusion}
\label{sec:conclusion}

Substituting one UD treebank for another reversed nearly 40\% of pairwise
language rankings for raw MDD, while every one of the 76 treebanks confirmed
dependency-length minimization (normalized ratio~$<$~1). The contrast
crystallizes MDD's dual character: the qualitative DLM universal survives
corpus substitution, but the ordinal cross-linguistic ranking does not. The
data favor MDD as a corpus-conditioned composite --- shaped by register,
annotation, and source characteristics alongside grammatical constraints ---
rather than a stable language-level parameter recoverable from any sufficiently
large sample. This does not invalidate MDD as a measure of syntactic processing
cost, but it shifts the burden of justification: studies treating a
single-corpus estimate as representative of a language must now argue that
corpus-specific factors are negligible for their comparison, rather than
assuming so by default.

%==================================================
% Acknowledgments

\section*{Acknowledgments}
None.

%==================================================
% Data and code availability

\section*{Data and Code Availability}

All source URLs, versions, licenses, checksums, inclusion decisions, derived
metrics, analysis code, and result objects are included in the reproduction
package. Raw corpus text is redistributed only where its license permits.

%==================================================
% References

\printbibliography

%==================================================
% Appendix

\section*{Appendix}

All supplementary tables referenced in the text, including the complete
inclusion and exclusion flow, treebank- and language-level estimates, sampling
stability curves, nested variance decomposition, specification-level results,
and machine-readable data files, are available in the online reproduction
package at \url{https://doi.org/10.5281/zenodo.20813136}.

\end{document}